\documentclass[10pt,twocolumn,letterpaper]{article}
\usepackage{cvpr}
\usepackage{times}
\usepackage{epsfig}
\usepackage{graphicx}
\usepackage{amsmath}
\usepackage{amssymb}
\usepackage{multirow}
\usepackage{subcaption}
\usepackage{booktabs}

\usepackage{amsmath,algorithm,algpseudocode,amsfonts,algorithmicx}
\newlength\savewidth
\newcommand\shline{\noalign{\global\savewidth\arrayrulewidth
                           \global\arrayrulewidth 1pt}%
                  \hline
                  \noalign{\global\arrayrulewidth\savewidth}}
                  
\newcommand\rhline{\noalign{\global\savewidth\arrayrulewidth
                           \global\arrayrulewidth 0.4pt}%
                  \hline
                  \noalign{\global\arrayrulewidth\savewidth}}
\usepackage[pagebackref=true,breaklinks=true,letterpaper=true,colorlinks,bookmarks=false]{hyperref}

\cvprfinalcopy % *** Uncomment this line for the final submission

\def\cvprPaperID{****} % *** Enter the CVPR Paper ID here

\begin{document}

%%%%%%%%% TITLE
\title{Learning Modulated Loss for Rotated Object Detection}

\author{Wen Qian$^{1,2}$ \quad {Xue Yang}$^{3}$ \quad {Silong Peng}$^{2}$ \quad {Yue Guo}$^{2}$ \quad {Junchi Yan}$^{3}$\\
${^1}$University of Chinese Academy of Sciences\\
${^2}$Institute of Automation, Chinese Academy of Sciences\\
${^3}$Shanghai Jiao Tong University \\
{\tt qianwen2018@ia.ac.cn \quad yangxue-2019-sjtu@sjtu.edu.cn
}
}

\maketitle

\begin{abstract}
     Popular rotated detection methods usually use five parameters (coordinates of the central point, width, height, and rotation angle) to describe the rotated bounding box and $\ell_1$ loss as the loss function. In this paper, we argue that the aforementioned integration can cause training instability and performance degeneration, due to the loss discontinuity resulted from the inherent periodicity of angles and the associated sudden exchange of width and height. This problem is further pronounced given the regression inconsistency among five parameters with different measurement units. We refer to the above issues as rotation sensitivity error (RSE) and propose a modulated rotation loss to dismiss the loss discontinuity. Our new loss is combined with the eight-parameter regression to further solve the problem of inconsistent parameter regression. Experiments show the state-of-art performances of our method on the public aerial image benchmark DOTA and UCAS-AOD. Its generalization abilities are also verified on ICDAR2015, HRSC2016, and FDDB. Qualitative improvements can be seen in Fig. \ref{fig:R_Qloss_vis}, and the source code will be released with the publication of the paper.
\end{abstract}

\section{Introduction}\label{sec:intro}
Object detection is an important and fundamental task in computer vision, and over the decades it has experienced a switch from traditional machine learning methods ~\cite{R1_viola2001rapid,R2_lienhart2002extended,R3_shotton2008multiscale,R4_lindeberg2012scale} to deep learning methods. The powerful fitting ability of the convolutional neural network~\cite{R5_Krizhevsky2012ImageNet} makes a series of great breakthroughs in the field of object detection, and performances of the neural network in many subdivision fields have surpassed those of human beings. Object detection has been extensively applied in face detection~\cite{R6_Zhang2017FaceBoxes,R7_Zhang2016Joint}, automatic driving~\cite{R8_Bagschik2017Ontology}, optical text detection~\cite{R9_Liao2016TextBoxes}, and other fields.

\begin{figure}[!tb]
    \centering
    \begin{subfigure}{.235\textwidth}
        \centering    \includegraphics[width=.98\linewidth,height=4cm]{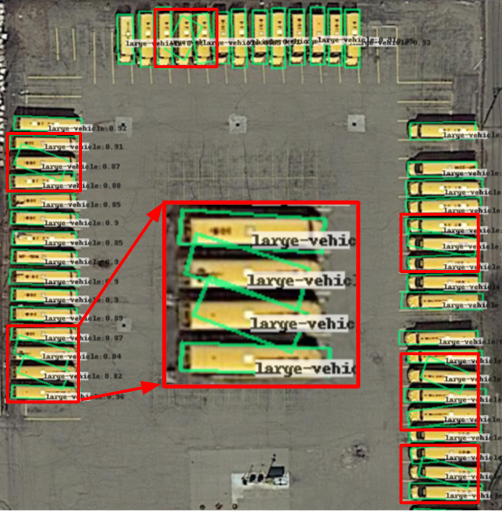}
        \caption{RetinaNet-H \cite{R20_Yang2019R3Det} (baseline)}
        \label{fig:R_Qloss_vis2}
    \end{subfigure}
    \begin{subfigure}{.235\textwidth}
        \centering    \includegraphics[width=.98\linewidth,height=4cm]{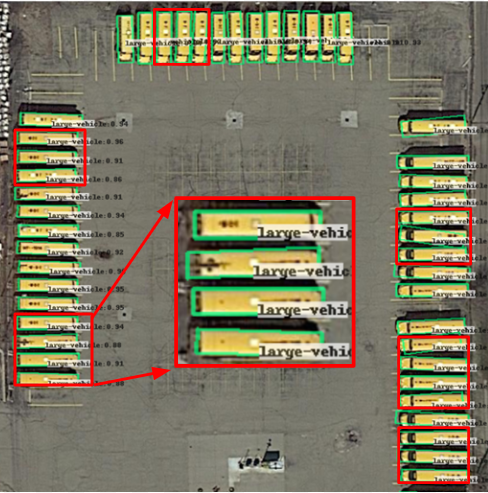}
        \caption{The proposed RSDet}
        \label{fig:R_Qloss_vis1}
    \end{subfigure}
    \vspace{-10pt}
    \caption{Detection results before and after solving the RSE problem with RSDet. The red rectangles in (a) represent failed examples due to the loss discontinuity.}
    \label{fig:R_Qloss_vis}
    \vspace{-15pt}
\end{figure}
\begin{figure}[h]
    \begin{center}
        \includegraphics[width=0.65\linewidth]{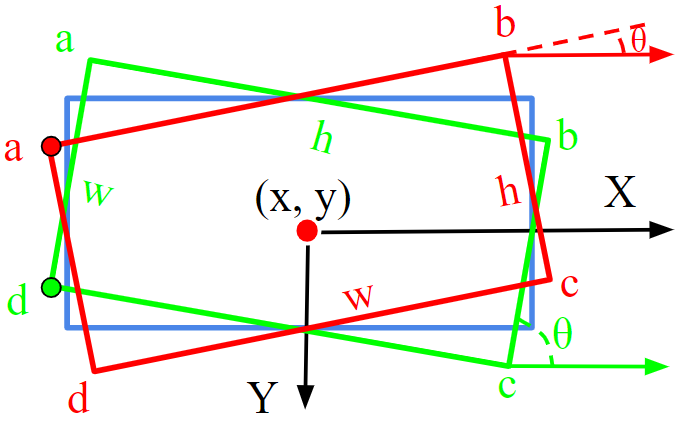}
    \end{center}
    \vspace{-10pt}
    \caption{Illustration for the loss discontinuity: rectangles colored in blue, red, and green respectively represent a reference box, a ground truth box, and a predicted box. Assume that the reference box is rotated one degree clockwise to get the ground truth one and is rotated similarly counterclockwise to obtain the predicted one. Consequently, the above three boxes are described with five parameters: the reference box (0, 0, 10, 25, -90$^{\circ}$), the ground truth box (0, 0, 25, 10, -1$^{\circ}$), and the predicted box (0, 0, 10, 25, -89$^{\circ}$). At this time, $\ell_1$ loss is far more than 0.}
    \label{fig:discontinuity_}
    \vspace{-10pt}
\end{figure}

Object detection can generally be divided into horizontal detection and rotation detection depending on directions of the detected boxes. Specifically, horizontal detection, by which all the bounding boxes are set in the horizontal direction, are often more suitable for general natural scene images such as COCO \cite{R23_lin2014microsoft} and Pascal VOC \cite{everingham2010pascal}. In contrast, in scene text, aerial imagery, face detection, and license plate detection, more accurate positioning is often needed and calls for effective rotation detectors. Until now, more and more rotated object detection datasets such as aerial dataset (DOTA \cite{R18_xia2018dota}, DIOR \cite{li2019object}, HRSC2016 \cite{R19_zk2017high}), scene text dataset (ICDAR2015 \cite{karatzas2015icdar}, ICDAR2017 \cite{gomez2017icdar2017}), and face dataset (FDDB \cite{jain2010fddb}) have appeared. The existing region-based rotated object detectors usually regress five parameters (coordinates of the central point, width, height, rotation angle) ~\cite{R20_Yang2019R3Det,R21_yang2018automatic,R22_Jiang2017R2CNN,R30_ma2018arbitrary} to describe rotated bounding boxes and use $\ell_1$-loss as loss functions. However, such methods bear two fundamental issues in practice:

Firstly, The loss discontinuity is caused by angle parameter. The loss value will jump when the angle reaches its range boundary, as shown in Fig. \ref{fig:discontinuity_}: a horizontal rectangle is respectively rotated one degree clockwise and counterclockwise to get the ground truth box and the detection box. The position of the reference rectangle has only been slightly changed, but its angle changes a lot due to parameterization. Moreover, the roles of height and width also exchange in a five-parameter system from OpenCV that makes the case more degenerating.

Moreover, in five-parameter system, parameters i.e. angle, width, height and center point have different measurement units, and show rather different relations against the Intersection over Union (IoU) (see Fig.~\ref{fig:inconsistency}). Simply adding them up for inconsistent regression can hurt performance. This issue may be mitigated by the eight-parameter system which use the coordinates of corners with the same units.

The above two issues, as referred collectively as rotation sensitivity error (RSE), which can lead to training instability (see Fig. ~\ref{fig:loss_curve}) and resulting detection performance degeneration. In order to address the loss discontinuity, a modulated rotation loss $\ell_{mr}$ is devised to carefully handle the boundary constraints for rotation, leading to a more smoothed loss curve during training. And then we resort to the eight-parameter regression model  \cite{liao2018rotation,Zhou2017EAST,zhang2019look,he2017deep} (As will be shown later in the paper, the discontinuity caused by rotation boundary inherently exist in eight-parameter system) to avoid regression parameter inconsistency. In the eight-parameter model, all the parameters are point coordinates of four corners in a bounding box such that the regression parameter consistency naturally holds.

In summary, we propose a rotation sensitive detector (RSDet) based on eight-parameter regression model and our modulated rotation loss $\ell_{mr}$, which can be trained end-to-end. Our RSDet model shows state-of-art performance on DOTA benchmark and its generalization capability and robustness are further verified on different datasets with different detectors. Our techniques are all orthogonal to existing methods. \textbf{The contributions of this paper are:}
%\begin{enumerate}
   % \item
   
    i) We formally formulate the important while relatively ignored rotation sensitivity error (RSE) for region-based rotation detectors, which refers to the loss discontinuity and regression inconsistency.
  %  \item
  
    ii) For the traditionally widely used five-parameter system (including the adoption in OpenCV), we formally identify the RSE in rotational object detectors. We then devise a special treatment to ensure the loss continuity. The new loss is termed by $\ell^{5p}_{mr}$ (see Eq.~\ref{eq:l_5pmr}).

 %   \item

    iii) Even for the more recent eight-parameter system, we still identify the inherent discontinuity issue and develop a corresponding treatment to smooth the loss function. The new loss is termed by $\ell^{8p}_{mr}$ (see Eq.~\ref{eq:l_8pmr}), and the resulting detector with RetinaNet-H \cite{Yang_2019_ICCV} as based model is termed rotation sensitive detector i.e. RSDet in this paper. 
%\end{enumerate}

\section{Related Work}
\paragraph{Horizontal Object Detectors}
Visual object detection has been a hot topic over the decade. Since the seminal work R-CNN \cite{girshick2014rich}, there have been a series of improvements including Fast RCNN \cite{girshick2015fast}, Faster RCNN \cite{R16_Ren2015Faster}, and R-FCN \cite{R15_dai2016r}, which fall the category of the two-stage methods. On the other hand, single-stage approaches have also been well developed which can be more efficient than the two-stage methods. Examples include Overfeat \cite{sermanet2013overfeat}, YOLO \cite{redmon2016you}, and SSD \cite{R11_liu2016ssd}. In particular, SSD \cite{R11_liu2016ssd} combines advantages of Faster RCNN and YOLO to achieve the trade-off between speed and accuracy. Subsequently, multi-scale feature fusion techniques are widely adopted in both single-stage methods and two-stage ones, such as FPN \cite{lin2017feature}, RetinaNet~\cite{R25_Lin2017Focal}, and DSSD~\cite{fu2017dssd}. Recently, many cascaded or refined detectors are proposed. For example, Cascade RCNN \cite{cai2018cascade}, HTC \cite{chen2019hybrid}, and FSCascade \cite{zhang2018single} perform multiple classifications and regressions in the second stage, leading to notable accuracy improvements in both localization and classification. Besides, the anchor free methods have become a new research focus, including FCOS~\cite{Tian_2019_ICCV}, FoveaBox~\cite{kong2019foveabox}, and RepPoints~\cite{Yang_2019_ICCV}. Structures of these detectors are simplified by discarding anchors, so anchor-free methods have opened up a new direction for object detection.

However, the above detectors only generate bounding boxes along the horizontal direction, which limits their applicability in many real-world scenarios. In fact, in scene texts and aerial images, objects tend to be densely arranged and have large aspect ratios, which requires more accurate localization. Therefore, rotated object detection has become a prominent direction in recent studies~\cite{Yang_2019_ICCV}.
\vspace{-10pt}
\paragraph{Rotated Object Detector}
Rotated object detection has been widely used in natural scene text, aerial image, etc. And these detectors typically use rotated bounding boxes to describe positions of objects, which are more accurate than those using horizontal boxes. Represented by scene text, many excellent detectors have been proposed. For example, RRPN \cite{R30_ma2018arbitrary} uses rotating anchors to improve the qualities of region proposals. R$^2$CNN \cite{R22_Jiang2017R2CNN} is a multi-tasking text detector that simultaneously detects rotated and horizontal bounding boxes. In TextBoxes++ \cite{R31_liao2018textboxes++}, to accommodate the slenderness of the text, a long convolution kernel is used and the number of proposals is increased. EAST \cite{Zhou2017EAST} proposes a simple yet powerful pipeline that yields fast and accurate text detection in natural scenes. 

Moreover, object detection in aerial images is more difficult, and its main challenges are reflected in multiple categories, multiple scales, complex backgrounds, dense arrangements, and a high proportion of small objects. Many scholars have also applied general object detection algorithms to aerial images, and many robust rotated detectors have emerged in aerial images. For example, ICN \cite{R27_azimi2018towards} combines various modules such as image pyramid, feature pyramid network, and deformable inception sub-networks, and it achieves satisfactory performances on DOTA dataset. RoI Transformer \cite{R29_ding2018learning} extracts rotation-invariant features for boosting subsequent classification and regression. SCRDet \cite{R28_Yang2018SCRDet} proposes an IoU-smooth $\ell_1$ loss to solve the sudden loss change caused by the angular periodicity so that it can better handle small, cluttered and rotated objects. R$^3$Det \cite{R20_Yang2019R3Det} proposes an end-to-end refined single-stage rotated object detector for fast and accurate object localization by solving the feature misalignment problem.

\begin{figure}[t]
    \centering
    \begin{subfigure}{.235\textwidth}
        \centering    \includegraphics[width=.98\linewidth]{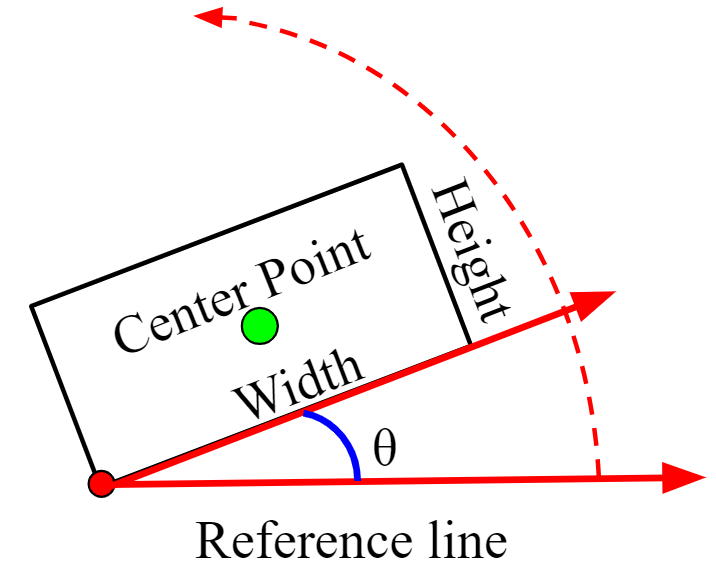}
        \caption{Width is longer than height.}
        \label{fig:3_1}
    \end{subfigure}
    \centering
    \begin{subfigure}{.235\textwidth}
        \centering    \includegraphics[width=.98\linewidth]{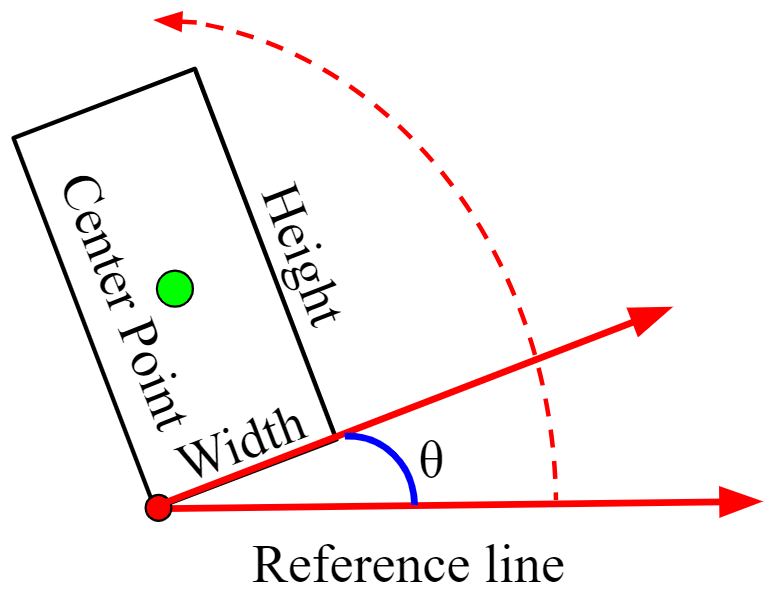}
        \caption{Height is longer than width.}
        \label{fig:3_2}
    \end{subfigure}
    \vspace{-10pt}
    \caption{The five-parameter definition in OpenCV exchanges the width and the height in the boundary condition for rotation. The angle parameter $\theta$ ranges from -90 degree to 0 degree, but it should be distinguished from another definition \cite{R18_xia2018dota}, with 180 degree angular range, whose $\theta$ is determined by the long side of the rectangle and x-axis.}
    \label{fig:define}
    \vspace{-15pt}
\end{figure}

All the above mentioned rotated object detectors do not consider the inherent loss discontinuity, as stated in Section~\ref{sec:intro}, which we show can hurt learning stability and final detection performances in our experiments. However, no existing studies have addressed this fundamental problem that motivates our work.
\vspace{-5pt}

%\textcolor{red}{Need some discussion on the limitation of these methods and the motivation of our works.}

\section{Proposed Method}
\label{sec:method}
\textbf{Overview.} In this section, we firstly present two mainstream protocols for bounding box parameterization \textit{i.e}. the five-parameter and eight-parameter models.
Then we formally identify the discontinuity of rotating angle and its resulting sudden exchange between width and height in the five-parameter system. Moreover, we quantitatively show the negative effect of the regression inconsistency in the five-parameter system caused by the different measurement units. We call such issues collectively as rotation sensitivity error (RSE) and propose a modulated rotation loss for the five-parameter system to achieve more smooth learning. We further point out that even the improved eight-parameter system still suffers from loss discontinuity and then devise a corresponding modulated rotation loss for the eight-parameter system.      

\begin{figure}[!tb]
    \centering
    \begin{subfigure}{.23\textwidth}
        \centering    \includegraphics[width=0.9\linewidth]{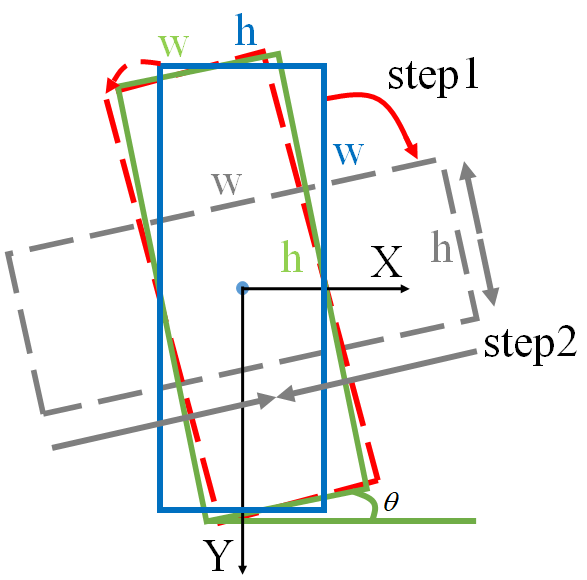}
        \caption{}
        \label{fig:ASE1}
    \end{subfigure}
    \begin{subfigure}{.23\textwidth}
        \centering    \includegraphics[width=0.9\linewidth]{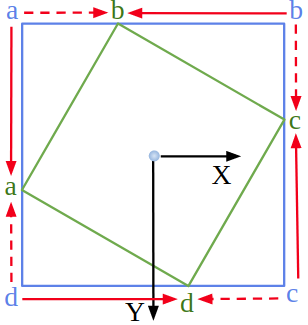}
        \caption{}
        \label{fig:ASE2}
    \end{subfigure}
    \vspace{-10pt}
    \caption{Boundary discontinuity analysis of five-parameter regression and eight-parameter regression. The red solid arrow indicates the actual regression process, and the red dotted arrow indicates the ideal regression process. (a) Five-parameter regression procedure including step 1 and step 2 under boundary conditions. (b) Eight-parameter regression procedure.}
    \label{fig:ASE}
    \vspace{-10pt}
\end{figure}

\subsection{Parameterization of Rotated Bounding Box}
Without loss of generality, our five-parameter definition is in line with that in OpenCV, as shown in Fig. \ref{fig:define}: a) define the reference line along the horizontal direction on which the vertex with the smallest vertical coordinate is located. b) rotate the reference line counterclockwise, the first rectangular side being touched by the reference line is defined as width $w$ regardless of its length compared with the other side -- height $h$. c) the central point coordinate is $(x, y)$ and the rotation angle is $\theta$.
% Note the popular five-parameter description in OpenCV can further exaggerate the discontinuity because width and height will exchange when the bounding box crosses the vertical line.

While the definition of eight parameters is more simple: four clockwise vertices (a,b,c,d) of the rotated bounding box are used to describe its position, as shown in Fig. ~\ref{fig:discontinuity_}. The eight-parameter regression methods have natural parameter consistency because it discards the inconsistency which is resulted from the non-coordinate parameter. At the same time, this kind of methods can describe quadrilateral, which can be used in more complex application scenarios.

\subsection{Rotation Sensitivity Error}
As mentioned earlier, rotation sensitivity error is mainly caused by two reasons: i) The adoption of angle parameter and the resulting height-width exchange (in the popular five-parameter description in OpenCV) contribute to the sudden loss change (increase) in the boundary case. ii) Regression inconsistency of measure units exists in the five-parameter model.
\begin{figure}[t]
    \centering
    \begin{subfigure}{.235\textwidth}
        \centering    \includegraphics[width=.98\linewidth]{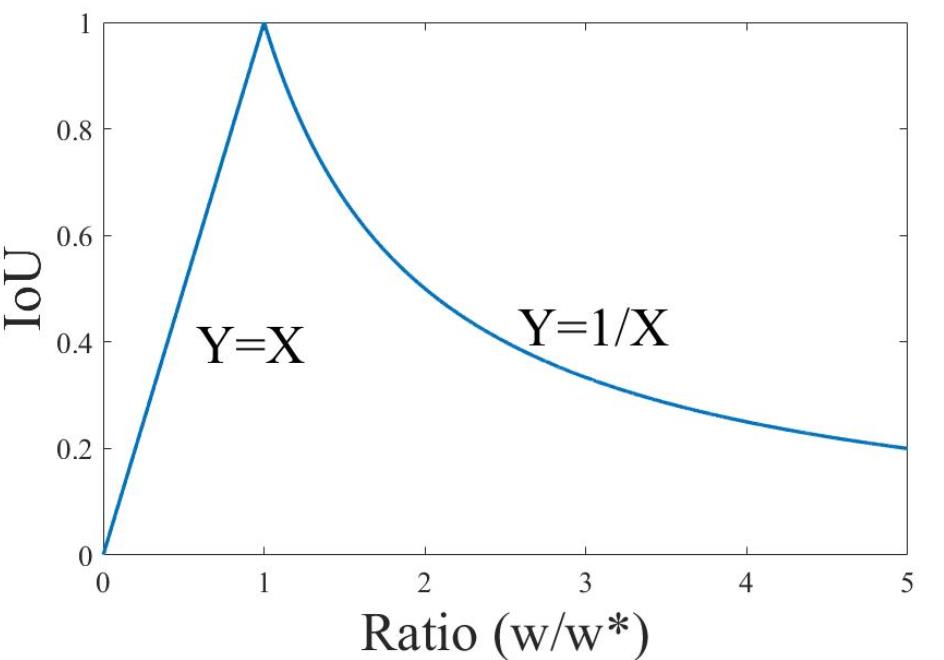}
        \vspace{-5pt}
        \caption{}
        \label{fig:2_1}
    \end{subfigure}
    \begin{subfigure}{.235\textwidth}
        \centering    \includegraphics[width=.98\linewidth]{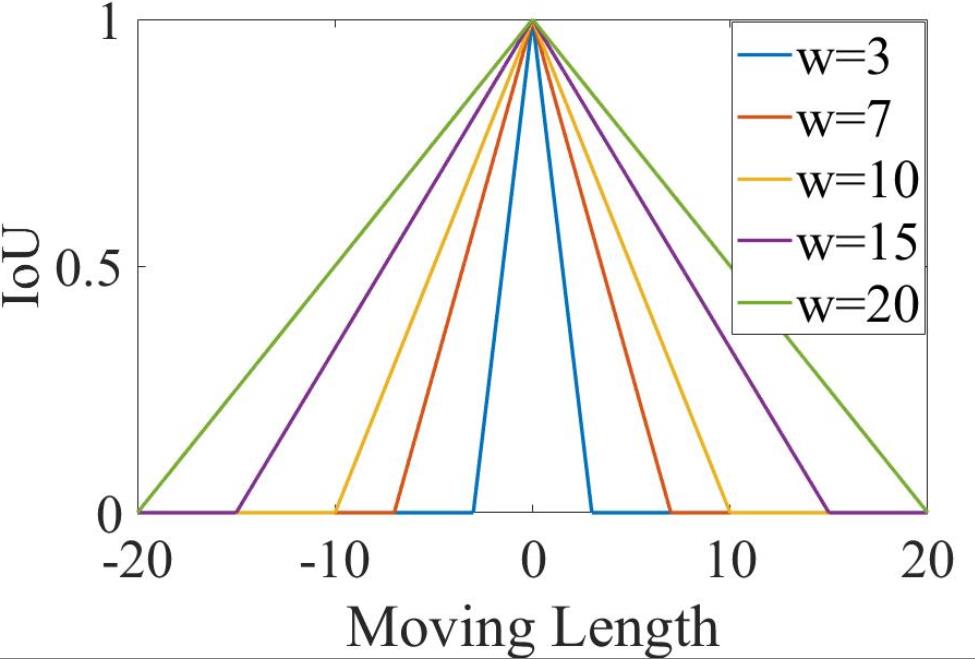}
        \vspace{-5pt}
        \caption{}
        \label{fig:2_2}
    \end{subfigure}
    \begin{subfigure}{.235\textwidth}
        \centering    \includegraphics[width=.98\linewidth]{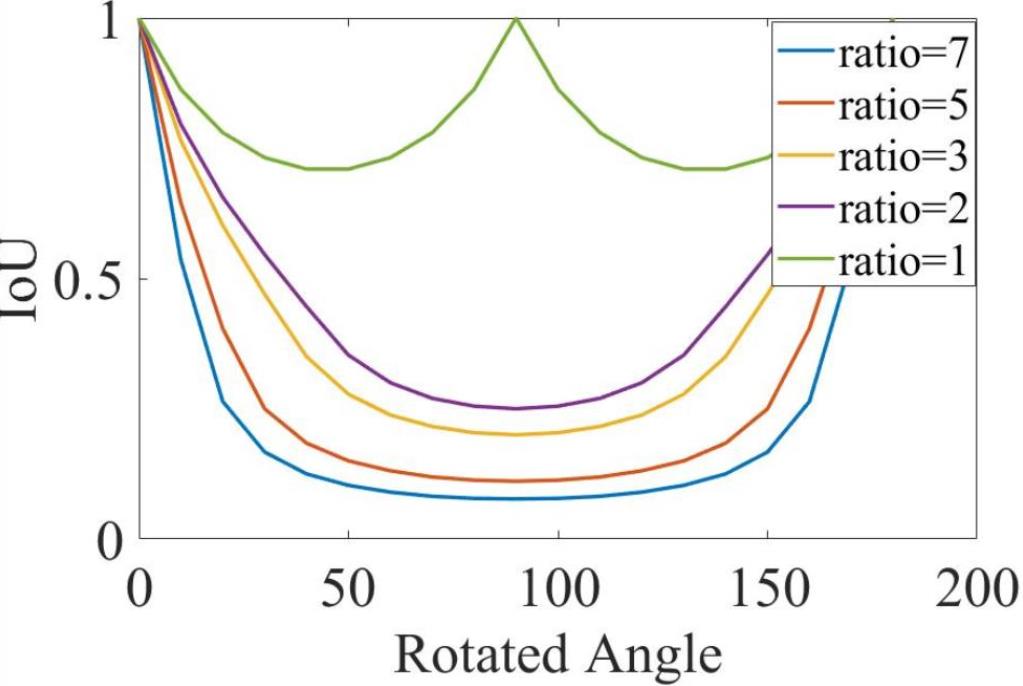}
        \vspace{-5pt}
        \caption{}
        \label{fig:2_3}
    \end{subfigure}
    \begin{subfigure}{.235\textwidth}
        \centering    \includegraphics[width=.98\linewidth]{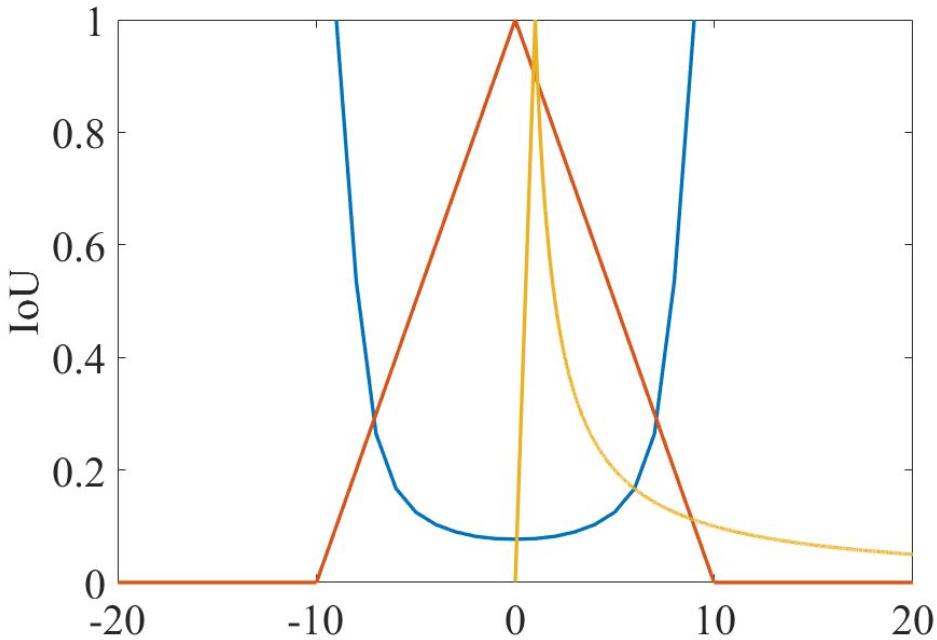}
        \vspace{-5pt}
        \caption{}
        \label{fig:2_4}
    \end{subfigure}
    \vspace{-10pt}
    \caption{Inconsistency in five-parameter regression model. (a) Relation between angle parameter and IoU. Different colors denote different aspect ratios. (b) Relation between width (similar for height) and IoU. (c) Relation between center point and IoU. (d) Comparison of three relations.}
    \label{fig:inconsistency}
\end{figure}

\begin{figure}[!tb]
    \centering
    \begin{subfigure}{.23\textwidth}
        \centering    \includegraphics[width=.98\linewidth]{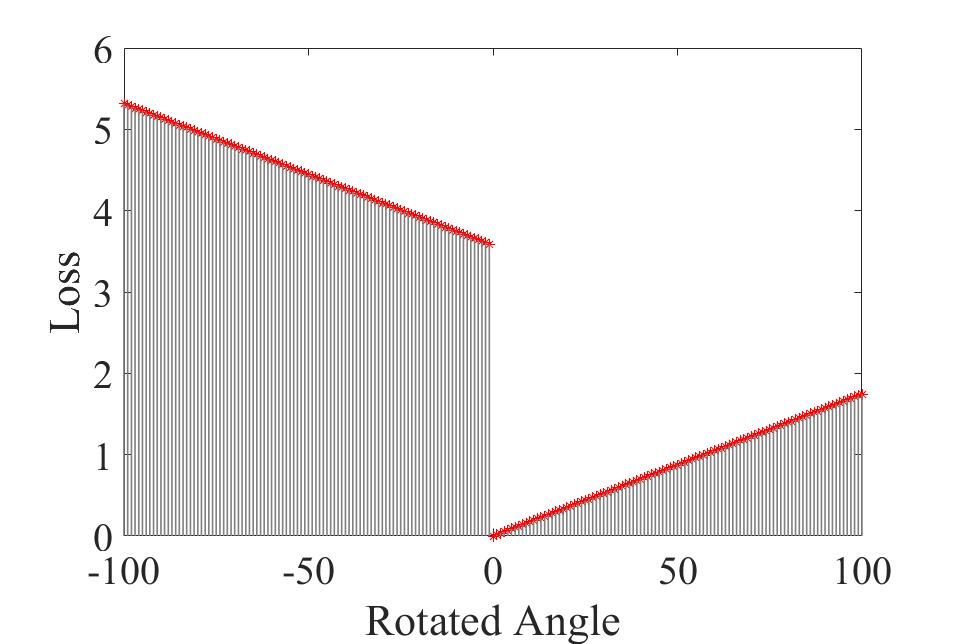}
        \caption{Discontinuous $\ell_1$-loss}
        \label{fig:fig4a}
    \end{subfigure}
    \begin{subfigure}{.23\textwidth}
        \centering    \includegraphics[width=.98\linewidth]{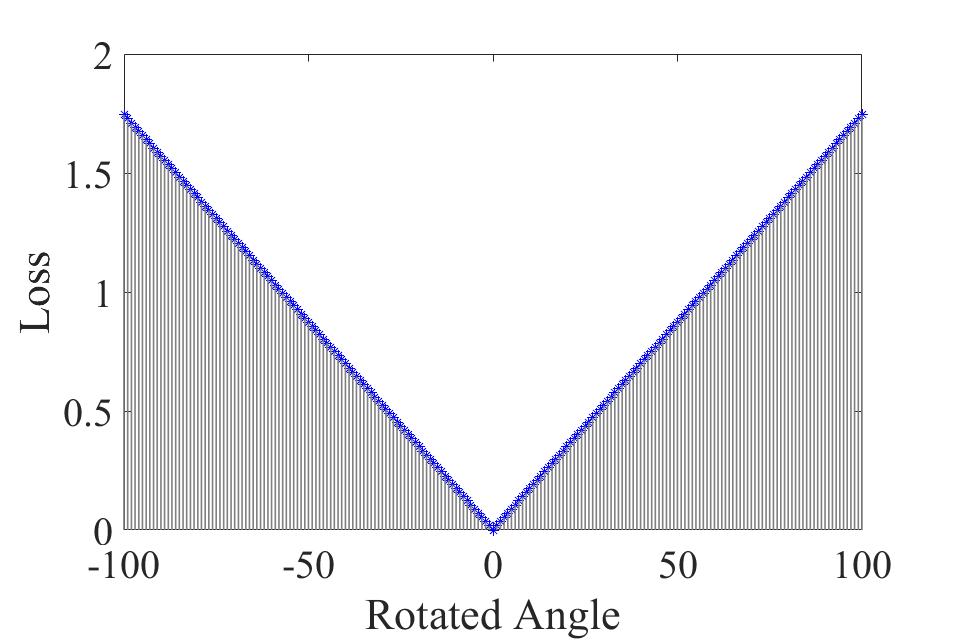}
        \caption{Continuous $\ell^{5p}_{mr}$}
        \label{fig:fig4b}
    \end{subfigure}
    \vspace{-10pt}
    \caption{Comparison between two loss functions.}
    \label{fig:discontinuity}
    \vspace{-10pt}
\end{figure}

\textbf{Loss Discontinuity.}
The angle parameter causes the loss discontinuous. To obtain the predicted box that coincides with the ground truth box, the horizontal reference box is rotated counterclockwise, as shown in Fig. \ref{fig:ASE}. In this figure, the coordinates of the predicted box are transformed from those of the reference box $(0,0,100,25,-90^\circ)$ to $(0,0,100,25,-100^\circ)$ in the normal coordinate system. However, the angle of the predicted box is out of the defined range, and the coordinates of the ground truth box are $(0,0,25,100,-10^\circ)$. Despite the rotation is physically smooth, the loss will be quite large, which corresponds to the loss discontinuity. To avoid such a loss fluctuation, the reference box need to be rotated clockwise to obtain the gray box $(0,0,100,25,-10^\circ)$ in Fig. \ref{fig:ASE1} (step 1), then width and height of the gray box will be scaled to obtain the final predicted box $(0,0,25,100,-10^\circ)$ (step 2). At this time, although the loss value is close to zero, the detector experiences a complex regression. This requires relatively high robustness, which increases the training difficulty. More importantly, an explicit and specific way is lacked to achieve a smooth regression, which will be addressed in the subsequent part of the paper.

\textbf{Regression Inconsistency.}
Different measurement units of five parameters make regression inconsistent. However, the impact of such artifacts is still unclear and has been rarely studied in the literature. Relationships among all the parameters and IoU are empirically studied in Fig. \ref{fig:inconsistency}. Specifically, the relationship between IoU and width (height) is a combination of a linear function and inverse proportion function, as illustrated in Fig. \ref{fig:2_3}. The relationship between the central point and IoU is a symmetric linear function, as illustrated in Fig. \ref{fig:2_2}. Completely different from other parameters, the relationship between the angle parameter and IoU is a multiple polynomial function (see Fig. \ref{fig:2_1}). Such regression inconsistency is highly likely to deteriorate the training convergence and the detection performance. Note that we use IoU as the standard measurement is because the final detection performance depends on whether IoU between the predicted box and the ground truth one is high enough.

%\subsection{The Proposed Modulated Rotation Loss}
\subsection{Five-parameter Modulated Rotation Loss}
 The loss discontinuity only occurs in the boundary case, as shown in Fig. ~\ref{fig:fig4a}. In this paper, we devise the following boundary constraints to modulate the loss as termed by modulated rotation loss $\ell_{mr}$:
\begin{equation}
    \begin{aligned}
        \ell_{cp}=|x_{1}-x_{2}|+|y_{1}-y_{2}|
    \end{aligned}
\end{equation}
\begin{equation}
    \begin{array}{l}
        \ell^{5p}_{mr}=\min\left\{\begin{array}{l}
            \ell_{cp}+|w_{1}-w_{2}|+|h_{1}-h_{2}| 
        +|\theta_{1}-\theta_{2}|
        \\           
        \ell_{cp}+|w_{1}-h_{2}|+|h_{1}-w_{2}| 
        +|90-|\theta_{1}-\theta_{2}||
        \end{array}\right.
    \end{array}
\end{equation}
where $\ell_{cp}$ is the central point loss. The first item in $\ell_{mr}$ is $\ell_1$-loss. The second item is a correction used to make the loss continuous by eliminating the angular periodicity and the exchangeability of height and width. This correction is particularly larger than $\ell_1$-loss when it does not reach the range boundary of the angle parameter. However, this correction becomes normal when $\ell_1$-loss is abrupt. In other words, such correction can be seen as the symmetry of $\ell_1$-loss about the location of the mutation. Finally, $\ell_{mr}$ takes the minimum of $\ell_1$-loss and the correction. The curve of $\ell_{mr}$ is continuous, as sketched in Fig. ~\ref{fig:fig4b}.

In practice, relative values of bounding box regression are usually used to avoid errors caused by objects on different scales. Therefore, $\ell_{mr}$ in this paper is expressed as follows:

\begin{equation}
    \begin{aligned}
    \nabla \ell_{cp}=|t_{x1}-t_{x2}|+|t_{y1}-t_{y2}|
    \end{aligned}
\end{equation}
    \begin{equation}
        \begin{aligned}
        \ell^{5p}_{mr}=\min\left\{\begin{array}{l}
        |t_{w1}-t_{w2}|+|t_{h1}-t_{h2}|+|t_{\theta 1}-t_{\theta 2}|+\nabla \ell_{cp}
        \\%[1mm]
        |t_{w1}-t_{h2}-\log(r)|+|t_{h1}-t_{w2}+\log(r)| \\%[1mm]
        \quad \quad \quad +||t_{\theta 1}-t_{\theta 2}| -\frac{\pi}{2}|+\nabla \ell_{cp}
        \end{array}\right.
        \end{aligned}
        \label{eq:l_5pmr}
    \end{equation}
    where
\begin{equation}
    \begin{aligned}
    \begin{array}{ll}
    t_{x}=(x-x_{a})/w_{a}, & t_{y}=(y-y_{a})/h_{a} \\%[1mm]
    t_{w}=\log(w/w_{a}), & t_{h}=\log(h/h_{a}) \\%[1mm]
    r=\frac{w}{h}, & t_{\theta}=\frac{\theta \pi}{180}
    \end{array}
    \end{aligned}
\end{equation}
where the measurement unit of the angle parameter is radian, $r$ represents the aspect ratio. $x$ and $x_{a}$ are respectively the predicted box and the anchor box (likewise for $y$, $w$, $h$, and $\theta$).

\subsection{Eight-parameter Modulated Rotation Loss}
%\textbf{Modulated rotation loss for eight-parameter model.} 
To avoid the inherent regression inconsistency, the eight-parameter representation has been recently developed~\cite{R31_liao2018textboxes++,liu2017deep,liu2019omnidirectional}. Specifically, the eight-parameter regression-based detectors directly regress the four corners of the object, so the prediction is a quadrilateral. The key step to the quadrilateral regression is to sort the four corner points in advance, which can avoid a very large loss even if the pose prediction is correct. For vertex order, we adopt a cross-product based algorithm to obtain the sequence of four vertices, as detailed in Algorithm \ref{alg:points_sequence}. Note that this algorithm is workable for convex quadrilaterals, and here we use the clockwise order for output without loss of generality. Algorithm \ref{alg:points_sequence} is similar to the counterpart proposed in Deep Matching Prior Network \cite{liu2017deep}.

However, the loss discontinuity still exists in the eight-parameter regression model. For example, we can suppose that a ground truth box can be described with the corner sequence $a\rightarrow b\rightarrow c\rightarrow d$ (see red box in Fig. \ref{fig:discontinuity_}. However, the corner sequence becomes $d\rightarrow a\rightarrow b\rightarrow c$ (see green box in Fig. \ref{fig:discontinuity_}) when the ground truth box is slightly rotated by a small angle. Therefore, consider the situation of an eight-parameter regression in the boundary case, as shown in Fig. \ref{fig:ASE2}. The actual regression process from the blue reference box to the green ground truth box is $\{({\color{blue}{a}}\rightarrow {\color{green}{a}}),({\color{blue}{b}}\rightarrow {\color{green}{b}}),({\color{blue}{c}}\rightarrow {\color{green}{c}}),({\color{blue}{d}}\rightarrow {\color{green}{d}})\}$, but apparently the ideal regression process should be $\{({\color{blue}{a}}\rightarrow {\color{green}{b}}),({\color{blue}{b}}\rightarrow {\color{green}{c}}),({\color{blue}{c}}\rightarrow {\color{green}{d}}),({\color{blue}{d}}\rightarrow {\color{green}{a}})\}$. This situation also causes the model training difficulty and the unsmooth regression.

Here we devise the eight-parameter version of our modulated rotation loss which consists of three components: i) move the four vertices of the predicted box clockwise by one place; ii) keep the order of the vertices of the predicted box unchanged; iii) move the four vertices of the predicted box counterclockwise by one place; iv) take the minimum value in the above three cases. Therefore, $\ell^{8p}_{mr}$ is expressed as follows:
    \begin{equation}
%\begin{footnotesize}
    \ell^{8p}_{mr}=\min\left\{\begin{split}
    &\sum_{i=0}^{3}\left(|x_{(i+3)\%4}-x^*_{i}|+|y_{(i+3)\%4}-y^*_{i}|\right)
    \\
    &\sum_{i=0}^{3}\left(|x_{i}-x^*_{i}|+|y_{i}-y^*_{i}|\right)
    \\
    &\sum_{i=0}^{3}\left(|x_{(i+1)\%4}-x^*_{i}|+|y_{(i+1)\%4}-y^*_{i}|\right)
    \end{split}\right.
    \label{eq:l_8pmr}
%\end{footnotesize}
    \end{equation}
where $x_{i}$ and $y_i$ respectively represent the coordinate offset between the $i$-th vertex of the predicted box and that of the reference box. $x^*_i$, $y^*_i$ respectively represents the offset between the $i$-th vertex of the ground truth box and that of the reference box.

\begin{algorithm}[tb!]
        \caption{Sequence ordering of quadrilateral corners.}
        \label{alg:points_sequence}
        \hspace*{0.02in} {\bf Input:}
        Four vertex of quadrilateral $\mathbf{p}_{1},\mathbf{p}_{2},\mathbf{p}_{3},\mathbf{p}_{4}$\\
        \hspace*{0.02in} {\bf Output:}
        Vertex in clockwise order: $\mathbf{p'}_{1},\mathbf{p'}_{2},\mathbf{p'}_{3},\mathbf{p'}_{4}$
        \begin{algorithmic}[1]
            \State $S \leftarrow \{\mathbf{p}_{1},\mathbf{p}_{2},\mathbf{p}_{3},\mathbf{p}_{4}\}$, $\mathbf{p'}_{2}=\mathbf{p'}_{3}=\mathbf{p'}_{4}=\mathbf{\textbf{0}}$;
            \State $\mathbf{p'}_{1} \leftarrow FindLeftmostVertex(S)$;
            \State $S \leftarrow S-\{\mathbf{p'}_{1}\}$
            
            \For{$\mathbf{s}_{1} \in S$}
            \State $\mathbf{s}_{2},\mathbf{s}_{3} \in S - \{\mathbf{s}_{1}\}$
            \If{$ CrossProduct(\mathbf{s}_{1}-\mathbf{p'}_{1}, \mathbf{s}_{2}-\mathbf{p'}_{1}) \times CrossProduct(\mathbf{s}_{1}-\mathbf{p'}_{1}, \mathbf{s}_{3}-\mathbf{p'}_{1}) < 0$}
            
            \State $\mathbf{p'}_{3}=\mathbf{s}_{1}$, $S \leftarrow \{\mathbf{s}_{2},\mathbf{s}_{3}\}$;
            \State \textbf{break};
            \EndIf
            \EndFor
            
            \For{$\mathbf{s}_{1} \in S$}
            \State $\mathbf{s}_{1}=S-\{\mathbf{s}_{1}\}$;
            \If{$CrossProduct(\mathbf{p'}_{3}-\mathbf{p'}_{1}, \mathbf{s}_{1}-\mathbf{p'}_{1}) > 0$}
            \State $\mathbf{p'}_{2}=\mathbf{s}_{1}$, $\mathbf{p'}_{4}=\mathbf{s}_{2}$;
            \Else
            \State $\mathbf{p'}_{2}=\mathbf{s}_{2}$, $\mathbf{p'}_{4}=\mathbf{s}_{1}$;
            \EndIf
            \EndFor
            \State \Return $\mathbf{p'}_{1},\mathbf{p'}_{2},\mathbf{p'}_{3},\mathbf{p'}_{4}$
        \end{algorithmic}
\end{algorithm}

\begin{figure}[!tb]
    \centering
    \begin{subfigure}{.235\textwidth}
        \centering    \includegraphics[width=.98\linewidth]{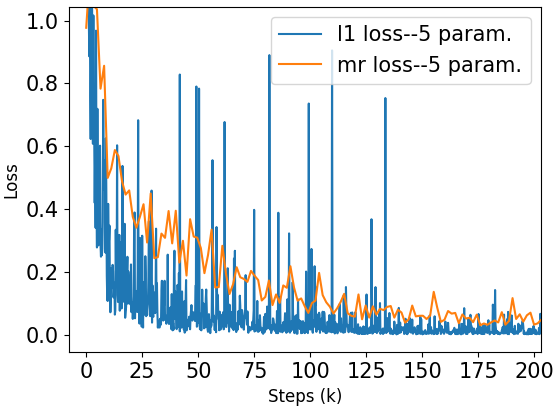}
        \caption{Loss curves (five-param.)}
        \label{fig:fig6_a}
    \end{subfigure}
    \begin{subfigure}{.235\textwidth}
        \centering    \includegraphics[width=.98\linewidth]{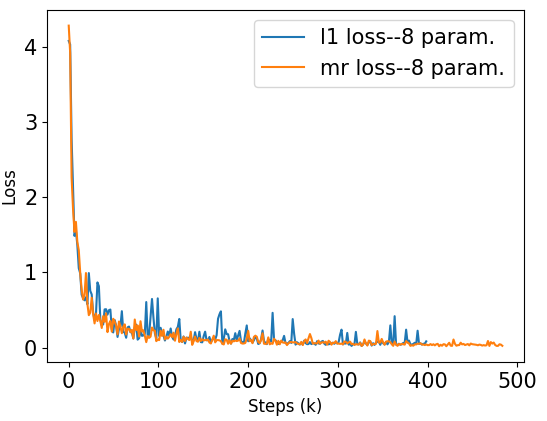}
        \caption{Loss curves (eight-param.)}
        \label{fig:fig6_b}
    \end{subfigure}
    \vspace{-10pt}
    \caption{Comparisons of loss curves during training with different loss functions.}
    \label{fig:loss_curve}
    \vspace{-10pt}
\end{figure}

Through the eight-parameter regression and the definition of $\ell^{8p}_{mr}$, the problems of the regression inconsistency and the loss discontinuity in rotation detection are eliminated. Extensive experiments show that our method is more stable for training (see Fig. \ref{fig:loss_curve}) and outperforms other methods.

\section{Experiments}
Recall that the main contribution of this paper is to identify the problem of RSE and solve it through modulated rotation loss and eight-parameter regression. Experiments are implemented by Tensorflow \cite{abadi2016tensorflow} on a server with Ubuntu 16.04, NVIDIA GTX 2080, and 32G Memory. Aerial images (DOTA and UCAS-AOD), scene text images (ICDAR2015 and HRSC2016), and face benchmark (FDDB) are used for evaluation.

\subsection{Datasets and Implementation Details}
\textbf{DOTA} \cite{R18_xia2018dota}: The main experiments are carried out around DOTA which has a total of 2,806 aerial images and 15 categories. The size of images in DOTA ranges from $800\times800$ pixels to $4,000\times4,000$ pixels. The proportions of the training set, the validation set, and the test set are respectively 1/2, 1/6, and 1/3. There are 188,282 instances for training and validation, and they are labeled with a clockwise quadrilateral. In this paper, we use the 1.0 version of annotations for rotated object detection. Due to the large size of a single aerial image, we divide the image into $600\times600$ pixel sub-images with a 150-pixel overlap between two neighboring ones, and these sub-images are eventually scaled to $800\times800$.

\textbf{ICDAR2015} \cite{karatzas2015icdar}: ICDAR2015 is a scene text dataset that includes a total of 1,500 images, 1000 of which are used for training and the remaining for testing. The size of the images in this dataset is $720\times1280$, and the source of the images is street view. The annotation of the text in an image is four clockwise point coordinates of a quadrangle.

\textbf{HRSC2016} \cite{R19_zk2017high}: HRSC2016 is a dataset for ship detection which range of aspect ratio and that of arbitrary orientation are large. This dataset contains two scenarios: ship on sea and ship close inshore. The size of each image ranges from $300\times300$ to $1,500\times900$. This dataset has 1061 images including 436 images for training, 181 images for validation, and 444 for testing.

% \textbf{FDDB} \cite{jain2010fddb}: FDDB is a dataset designed for unconstrained face detection, in which faces have a wide variability of face scales, poses, and appearance. This dataset contains annotations for 5,171 faces in a set of 2,845 images taken from the faces in the Wild dataset~\cite{berg2005s}.

\textbf{UCAS-AOD} \cite{zhu2015orientation}: UCAS-AOD is a remote sensing dataset which contains two categories: car and plane. UCAS-AOD contains 1510 aerial images, each of which has approximately $659\times1,280$ pixels. In line with \cite{R29_ding2018learning} and \cite{R27_azimi2018towards}, we randomly select 1110 images for training and 400 ones for test.

\textbf{Baselines and Training Details.}
To make the experimental results more reliable, the baseline we chose is a multi-class rotated object detector based on RetinaNet, which has been verified in work \cite{R20_Yang2019R3Det}. During training, we use RetinaNet-Res50, RetinaNet-Res101, and RetinaNet-Res152 ~\cite{R25_Lin2017Focal} for experiments. Our network is initialized with the pre-trained ResNet50 ~\cite{he2016deep} for object classification in ImageNet \cite{deng2009imagenet}, and the pre-trained models are officially published by TensorFlow. Besides, weight decay and momentum are correspondingly 1e-4 and 0.9. The training epoch is 30 in total, and the number of iterations per epoch depends on the number of samples in the dataset. The initial learning rate is 5e-4, and the learning rate changes from 5e-5 at epoch 18 to 5e-6 at epoch 24. In the first quarter of the training epochs, we adopt the warm-up strategy to find a suitable learning rate. During the inference, rotating non-maximum suppression (R-NMS) is used for post-processing the final detection results.

\begin{table}[!tp]
    \centering
    \resizebox{0.40\textwidth}{!}{
        \begin{tabular}{l|lc|c}
            % \hline
            Backbone & Loss & Regression & mAP\\
            \shline
             resnet-50 & smooth-$\ell_1$ & five-param. & 62.14\\
             resnet-50 & $\ell_{mr}$ & five-param. & 64.49\\
             resnet-50 & smooth-$\ell_1$ & eight-param. & 65.59\\
             resnet-50 & $\ell_{mr}$ & eight-param. & \textbf{66.77}\\
            % \hline
        \end{tabular}}
        \vspace{-5pt}
        \caption{Ablation experiments of $\ell_{mr}$ and predefined eight-parameter regression on DOTA benchmark. RetinaNet-H \cite{R20_Yang2019R3Det} is used as the baseline.}
        \label{tab:ablation_dota}
        \vspace{-5pt}
\end{table}

\begin{table}[!tp]
    \centering
    \resizebox{0.40\textwidth}{!}{
    \begin{tabular}{l|l|c}
        Loss & Regression & mAP\\
        \shline
        smooth-$\ell_1$ & five-param. [$-\frac{\pi}{2}$,0) & 62.14\\
        smooth-$\ell_1$ & five-param. [$-\pi$,0) \cite{R18_xia2018dota} & 62.39\\
        smooth-$\ell_1$ & five-param. [$-\frac{\pi}{2}$,0)+tan \cite{R30_ma2018arbitrary, bao2019single}  & NAN\\
        IoU-smooth-$\ell_1$ \cite{R28_Yang2018SCRDet} & five-param. [$-\frac{\pi}{2}$,0) & 62.69\\
        \rhline
        $\ell_{mr}$ & five-param. [$-\frac{\pi}{2}$,0) & 64.49 \\
        smooth-$\ell_1$ & eight-param. & 65.59\\
        $\ell_{mr}$ & eight-param. & \textbf{66.77}\\
    \end{tabular}}
    \vspace{-5pt}
    \caption{Ablation study using the proposed techniques on DOTA. RetinaNet-H\cite{R20_Yang2019R3Det} is the base model.}
    \label{tab:similar_methods}
    \vspace{-10pt}
\end{table}

\begin{table*}[h]
    \centering
    \resizebox{0.95\textwidth}{!}{
    \begin{tabular}{l|ccccccccccccccc|c}
        
        Method&PL&BD&BR&GTF&SV&LV&SH&TC&BC&ST&SBF&RA&HA&SP&HC&mAP\\
        \shline
        FR-O ~\cite{R18_xia2018dota} & 79.1 & 69.1 & 17.2 & 63.5 & 34.2 & 37.2 & 36.2 & 89.2 & 69.6 & 59.0 & 49.4 & 52.5 & 46.7 & 44.8 & 46.3 & 52.9\\
        R$^2$CNN~\cite{R22_Jiang2017R2CNN} & 80.9 & 65.7 & 35.3 & 67.4 & 59.9 & 50.9 & 55.8 & 90.7 & 66.9 & 72.4 & 55.1 & 52.2 & 55.1 & 53.4 & 48.2 & 60.7\\
        RRPN~\cite{R30_ma2018arbitrary} & 88.5 & 71.2 & 31.7 & 59.3 & 51.9 & 56.2 & 57.3 & 90.8 & 72.8 & 67.4 & 56.7 & 52.8 & 53.1 & 51.9 & 53.6 & 61.0\\
        
        RetinaNet-H+ResNet50 \cite{R20_Yang2019R3Det} & 88.9 & 74.5 & 40.1 & 58.0 & 63.1 & 50.6 & 63.6 & 90.9 & 77.9 & 76.4 & 48.3 & 55.9 & 50.7 & 60.2 & 34.2 & 62.2 \\
        
        RetinaNet-R+ResNet50 \cite{R20_Yang2019R3Det} & 88.9 & 67.7 & 33.6 & 56.8 & 66.1 & 73.3 & 75.2 & 90.9 & 74.0 & 75.1 & 43.8 & 56.7 & 51.1 & 55.7 & 21.5 & 62.0 \\
        
        ICN~\cite{R27_azimi2018towards} & 81.4 & 74.3 & 47.7 & 70.3 & 64.9 & 67.8 & 70.0 & 90.8 & 79.1 & 78.2 & 53.6 & 62.9 & 67.0 & 64.2 & 50.2 & 68.2\\
        RoI Transformer~\cite{R29_ding2018learning} & 88.6 & 78.5 & 43.4 & \textbf{75.9} & 68.8 & 73.7 & 83.6 & 90.7 & 77.3 & 81.5 & 58.4 & 53.5 & 62.8 & 58.9 & 47.7 &  69.6\\
        SCRDet~\cite{R28_Yang2018SCRDet} & 90.0 & 80.7 & 52.1 & 68.4 & 68.4 & 60.3 & 72.4 & 90.9 & 88.0 & \textbf{86.9} & 65.0 & 66.7 & 66.3 & 68.2 & 65.2 & 72.6 \\
        
        R$^3$Det+ResNet152 \cite{R20_Yang2019R3Det} & 89.2 & 80.8 & 51.1 & 65.6 & \textbf{70.7} & 76.0 & \textbf{78.3} & 90.8 & 84.9 & 84.4 & \textbf{65.1} & 57.2 & \textbf{68.1} & 69.0 & 60.9 & 72.8\\
        \rhline
        RSDet+ResNet50 (ours) & 89.3 & 82.7 & 47.7 & 63.9 & 66.8 & 62.0 & 67.3 & 90.8 & 85.3 & 82.4 & 62.3 & 62.4 & 65.7 & 68.6 & 64.6 & 70.8\\
        RSDet+ResNet101 (ours) & 89.8 & 82.9 & 48.6 & 65.2 & 69.5 & 70.1 & 70.2 & 90.5 & 85.6 & 83.4 & 62.5 & 63.9 & 65.6 & 67.2 & \textbf{68.0} & 72.2\\
        RSDet+ResNet152 (ours) & \textbf{90.2} & \textbf{83.5} & 53.6 & 70.1 & 64.6 & 79.4 & 67.3 & 91.0 & \textbf{88.3} & 82.5 & 64.1 & \textbf{68.7} & 62.8 & 69.5 & 66.9 & 73.5\\
        RSDet+ResNet152+Refine
        (ours) & 90.1 & 82.0 & \textbf{53.8} & 68.5 & 70.2 & \textbf{78.7} & 73.6 & \textbf{91.2} & 87.1 & 84.7 & 64.3 & 68.2 & 66.1 & \textbf{69.3} & 63.7 & \textbf{74.1}\\
    \end{tabular}}
    \caption{Detection accuracy (AP for each category and overall mAP) on different objects and overall performances with the state-of-the-art methods on DOTA. The short names for categories are defined as (abbreviation-full name): PL-Plane, BD-Baseball diamond, BR-Bridge, GTF-Ground field track, SV-Small vehicle, LV-Large vehicle, SH-Ship, TC-Tennis court, BC-Basketball court, ST-Storage tank, SBF-Soccer-ball field, RA-Roundabout, HA-Harbor, SP-Swimming pool, and HC-Helicopter. For RetinaNet, 'H' and 'R' denote horizontal anchors and the rotated anchors, respectively.}
    \label{tab:dota_sota}
    \vspace{-8pt}
\end{table*}

\subsection{Ablation Study}
\textbf{Modulated Rotation Loss and Eight-parameter Regression.}
We use the ResNet50-based RetinaNet-H as our baseline to verify the effectiveness of modulated rotation loss $\ell_{mr}$ and eight-parameter regression. We get a gain of 2.35\% mAP, when the loss function is changed from the smooth-$\ell_1$ loss to $\ell_{mr}$, as shown in Tab.~\ref{tab:ablation_dota}. Fig. \ref{fig:R_Qloss_vis} compares results before and after solving the RSE problem: objects in the images are all in the boundary case where the loss function is not continuous. A lot of inaccurate results (see red circles in Fig. \ref{fig:R_Qloss_vis2}) are predicted in the baseline method, but these do not occur after using $\ell_{mr}$ (see the same location in Fig. \ref{fig:R_Qloss_vis1}). Similarly, an improvement of 3.45\% mAP is obtained after using the eight-parameter regression. Finally, we achieve 66.77\% mAP after combining these two techniques. This set of ablation experiments prove that $\ell_{mr}$ and eight-parameter regression are effective for improving the rotated object detector. More importantly, the number of parameters and calculations added by these two techniques are almost negligible.

\textbf{Training Stability.}
In Section \ref{sec:method}, we have analyzed that the loss discontinuity and the regression inconsistency greatly affect the training stability and the detection performance in detail. Although the detection performance using our techniques has been verified through mAPs, we have not proven the stability improvement of model training brought by our techniques. To this end, we plot the training loss curves using models including RetinaNet-H ($\ell^{5p}$), RetinaNet-H ($\ell^{5p}_{mr}$), RetinaNet-H ($\ell^{8p}$), and RSDet ($\ell^{8p}_{mr}$), as shown in Fig. \ref{fig:loss_curve}. We can see that training convergences become more stable after using modulated rotation losses.

\begin{table}[!tp]
%    \small
    \centering
    \resizebox{0.4\textwidth}{!}{
    \begin{tabular}{l|cc|c}
        % \hline
         Backbone&Data Augmentation&Balance&mAP\\
        \shline
        resnet-50 &  &  & 66.77\\
         resnet-50 & $\checkmark$ &  & 70.79\\
         resnet-50 & $\checkmark$ & $\checkmark$ & 71.22\\
         resnet-101 & $\checkmark$ & $\checkmark$ & 72.16\\
         resnet-152 & $\checkmark$ & $\checkmark$ & \textbf{73.51}\\
        % \hline
    \end{tabular}}
    \vspace{-5pt}
    \caption{Ablation experiments of backbone, data augmentation and balance on DOTA. RSDet is the base model.}
    \label{tab:data_backbone}
    \vspace{-10pt}
\end{table}

\textbf{Comparison with Similar Methods.}
Although we formally introduce the concept of RSE for the first time, it is worth noting that some previous articles have also mentioned similar problems. In \cite{R18_xia2018dota}, a 180-degree angle definition is used to eliminate the loss burst caused by the exchangeability of height and width. While related works \cite{R30_ma2018arbitrary, bao2019single} use periodic trigonometric functions (such as $tan$) to eliminate the effects of the angular periodicity. SCRDet \cite{R28_Yang2018SCRDet} proposes IoU-smooth-$\ell_1$ loss to solve the boundary discontinuity. However, these methods are limited and do not completely solve the RSE problem. Tab. \ref{tab:similar_methods} compares our proposed method with other methods mentioned above. Our approach still yields the most promising results.

\begin{figure*}[t]
    \centering
    \begin{subfigure}{.23\textwidth}
        \centering    
        \includegraphics[width=0.98\linewidth, height=3.4cm]{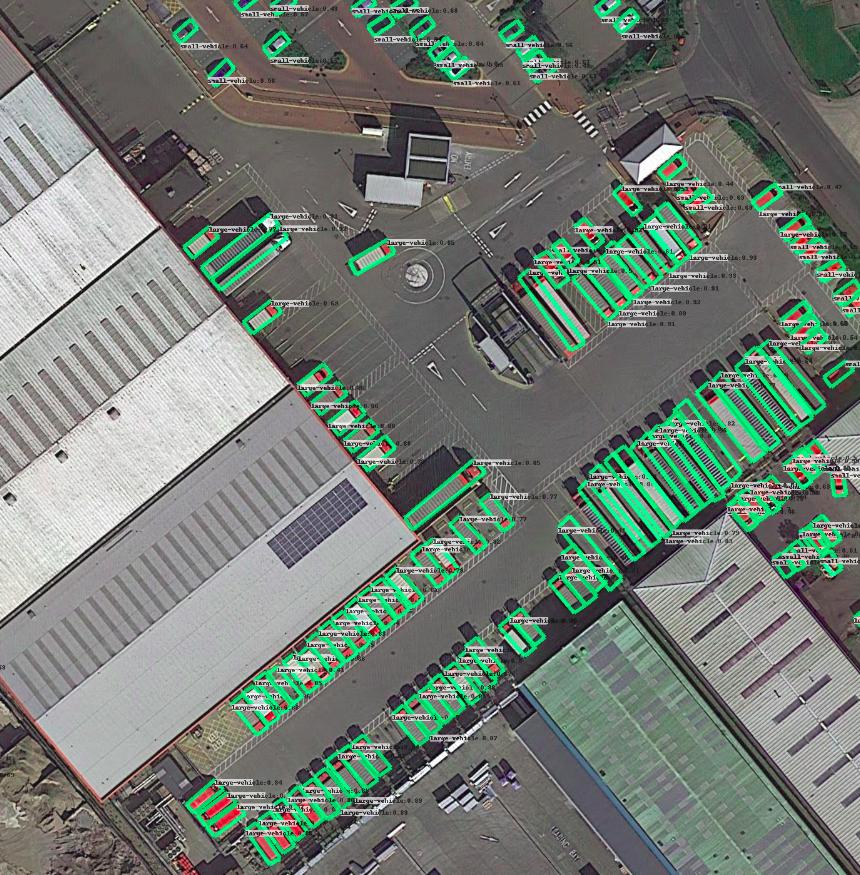}
        \caption{Vehicles}
    \end{subfigure}
    \begin{subfigure}{.23\textwidth}
        \centering            
        \includegraphics[width=0.98\linewidth, height=3.4cm]{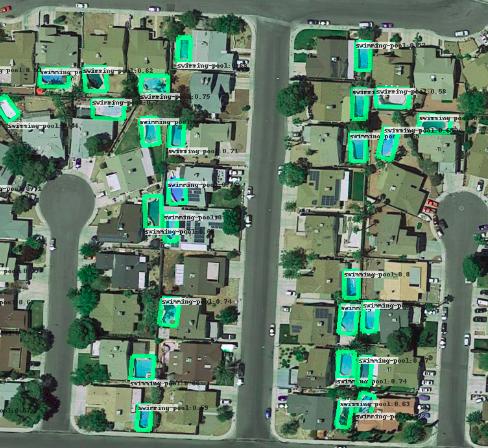}
        \caption{Swimming pool}
    \end{subfigure}
    \begin{subfigure}{.23\textwidth}
        \centering    
        \includegraphics[width=0.98\linewidth, height=3.4cm]{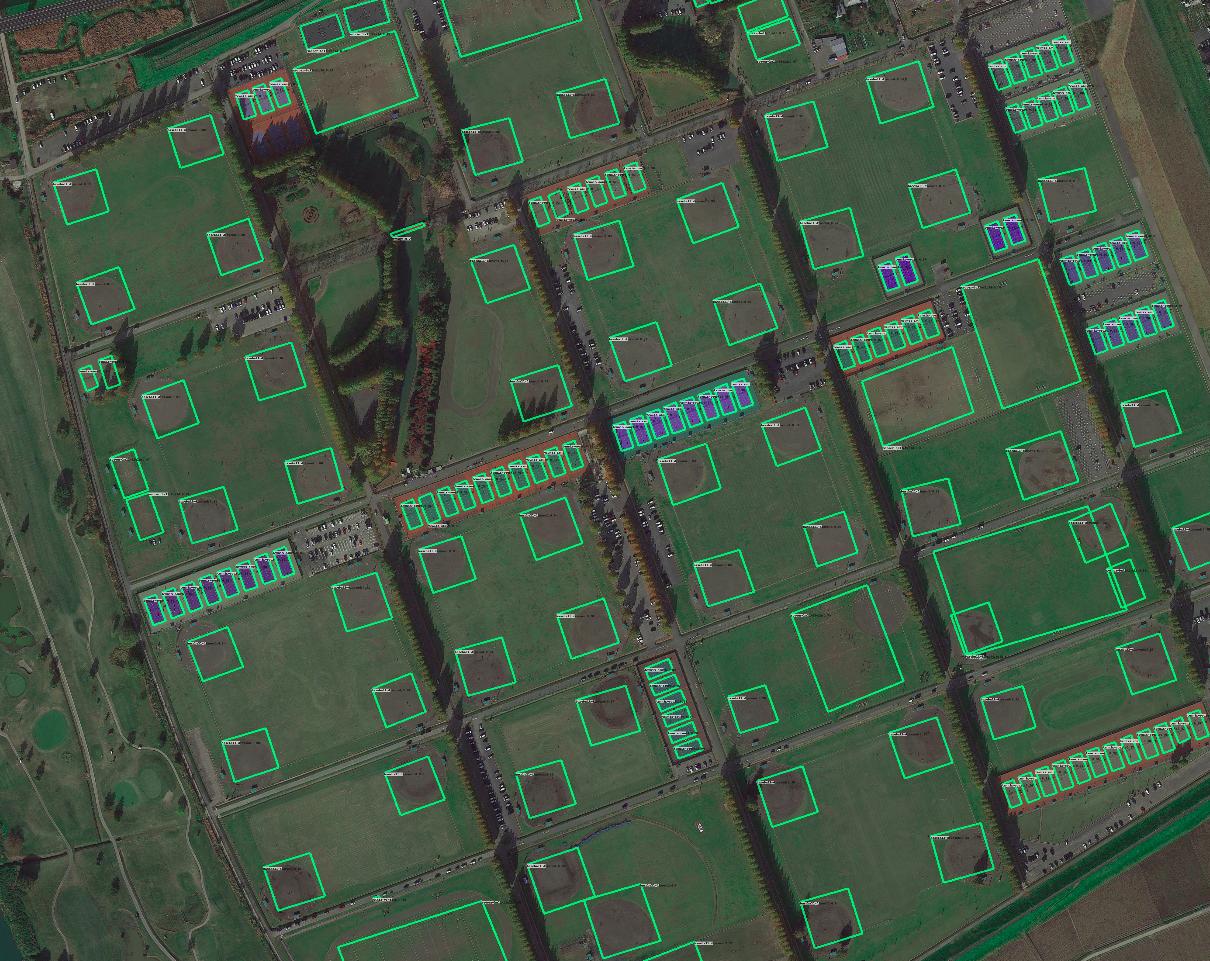}
        \caption{Tennis and soccer-ball field}
    \end{subfigure}
    \begin{subfigure}{.23\textwidth}
        \centering    
        \includegraphics[width=0.98\linewidth, height=3.4cm]{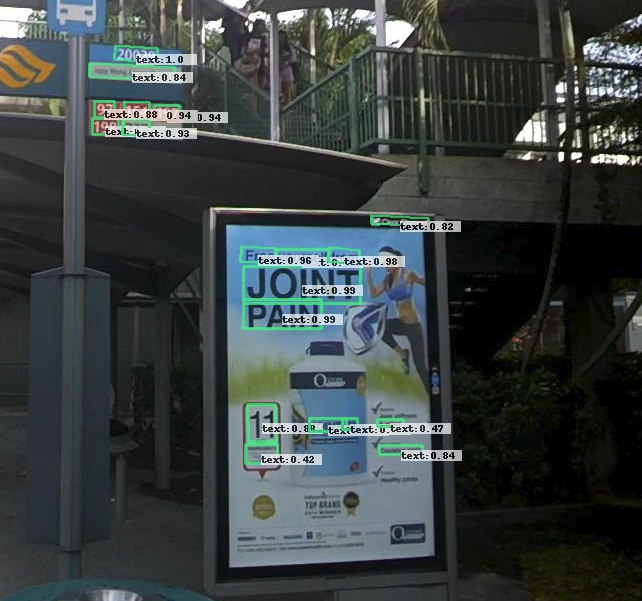}
        \caption{Words on bulletin board}
    \end{subfigure}
    \\
    \begin{subfigure}{.23\textwidth}
        \centering    
        \includegraphics[width=0.98\linewidth, height=3.4cm]{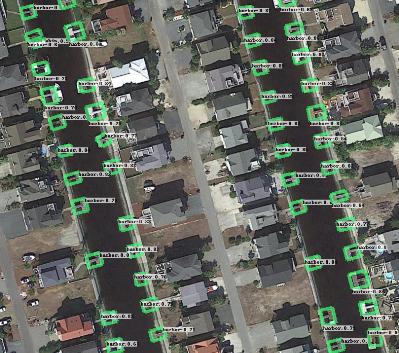}
        \caption{Harbor}
    \end{subfigure}    
    \begin{subfigure}{.23\textwidth}
        \centering    
        \includegraphics[width=0.98\linewidth, height=3.4cm]{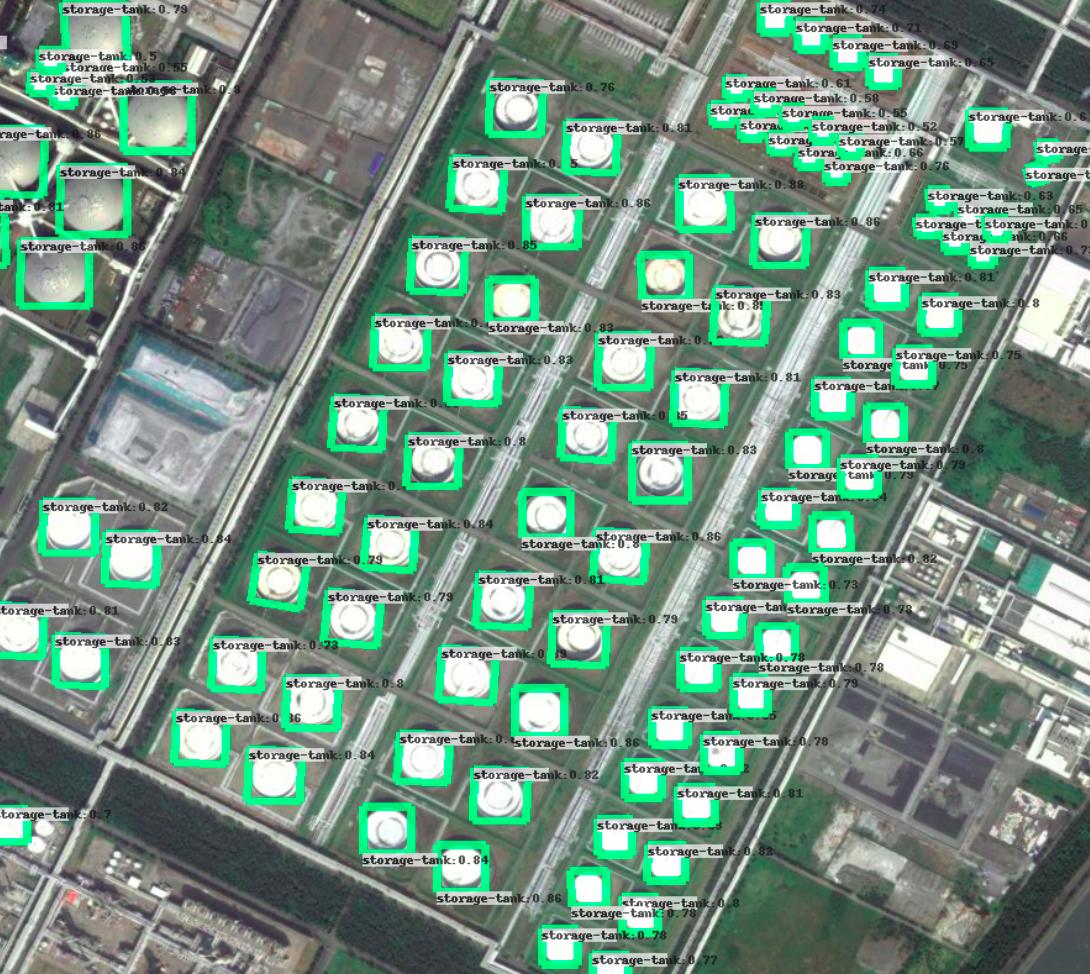}
        \caption{Storage tank}
    \end{subfigure}
    \begin{subfigure}{.23\textwidth}
        \centering    
        \includegraphics[width=0.98\linewidth, height=3.4cm]{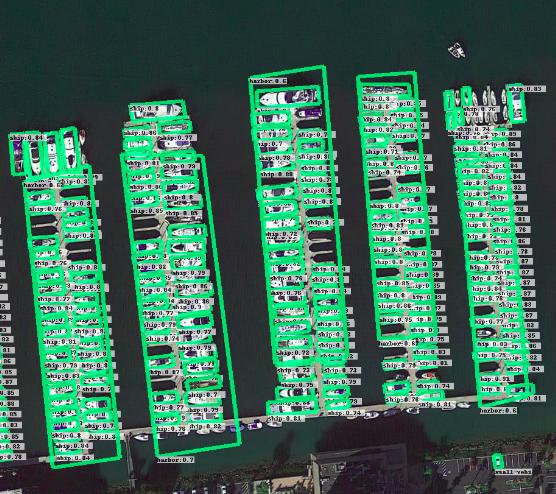}
        \caption{Harbor and ship}
    \end{subfigure}
   \begin{subfigure}{.23\textwidth}
        \centering            
        \includegraphics[width=0.98\linewidth, height=3.4cm]{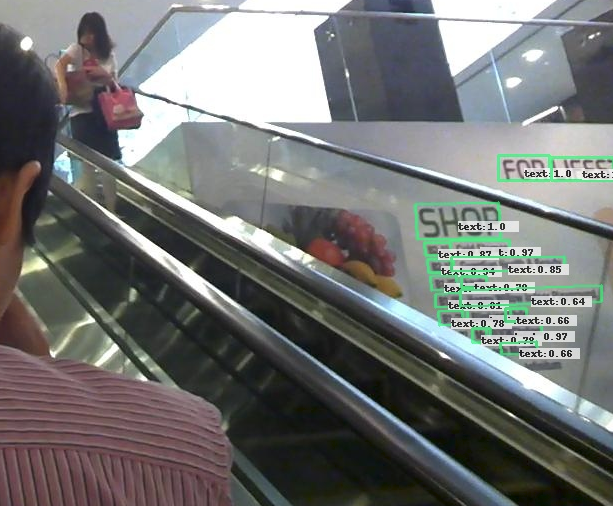}
        \caption{Text seen on the elevator}
    \end{subfigure}
    \vspace{-5pt}
    \caption{Detection results on DOTA and ICDAR15.}
    \label{fig:dota_vis}
    \vspace{-8pt}
\end{figure*}

\textbf{Backbone, Data Augmentation, and Data Balance.}
Data augmentation is effective to improve detection performance. Operations of augmentations we use include random horizontal flipping, random vertical flipping, random image graying, and random rotation. Consequently, the baseline performance increased by 4.22\% to 70.79\% on DOTA. Data imbalance is severe in the DOTA. For instance, there are 76,833 ship instances in the dataset, but there are only 962 ground track fields. We extend samples fewer than 10,000 to 10,000 ones in each category by copying, which brings a 0.43\% boost, and the most prominent contribution is from a small number of samples such as helicopter and swimming pool. We also explore the impact of different backbones on the detector and conclude that larger backbones bring more performance gains. Performances of the detectors based on ResNet50, ResNet101, and ResNet152 are respectively 71.22\%, 72.16\% and 73.51\%. Refer to Tab. \ref{tab:data_backbone} for detailed results.

\begin{table}[!tp]
    \centering
    \resizebox{0.4\textwidth}{!}{
    \begin{tabular}{l|lc|c}
         Loss & Regression& ICDAR2015 & HRSC2016\\
        \shline
         smooth-$\ell_1$ & five-param. & 76.8 & 82.4\\
         $\ell_{mr}$ & five-param. & 79.6 & 83.6\\
          smooth-$\ell_1$ & eight-param. & 81.2 & 85.4\\
         $\ell_{mr}$ & eight-param. & \textbf{83.2} & \textbf{86.5}\\
    \end{tabular}}
    \vspace{-5pt}
    \caption{Performances of $\ell_{mr}$ and eight-parameter regression on ICDAR2015 and HRSC2016. RetinaNet-H \cite{R20_Yang2019R3Det} is the base model, and ResNet152 is the backbone.}
    \label{tab:hrsc2016_icdar2015}
    \vspace{-10pt}
\end{table}

\textbf{Regression Refinement.}
R$^{3}$Det proposes to increase the accuracy of the regression box by adding a refinement stage. The idea of cascading is also used by other detection methods, such as RefineDet \cite{zhang2018single} and Cascade RCNN \cite{cai2018cascade}. Therefore, in order to further improve the performance of RSDet, we also add a refinement stage. The foreground and background thresholds for the refinement stage are 0.6 and 0.5, respectively. Tab.~\ref{tab:dota_sota} shows that the refinement stage helps improve the performance from 73.51\% to 74.13\%, especially for objects with large aspect ratios such as ships, vehicles, and harbors.

\textbf{Using Two-stage Detectors as Base Model.}
Extra experiments are performed based on the Rotating Faster RCNN. Unlike RetinaNet, Faster RCNN is a two-stage detector. We take a rotating Faster RCNN as the baseline, then add the $\ell_{mr}$ and eight-parameter regression method for the ablation experiments. The performance improvement of these two techniques are 1.6\% and 2.84\%, respectively.

% \begin{figure}[!tb]
%     \centering
    % \begin{subfigure}{.235\textwidth}
    %     \centering    
    %     \includegraphics[width=.98\linewidth, height=3.2cm]{figure/vis_dota/face1.jpg}
    %     \caption{Face in grayscale image}
    %     \label{fig:fig9_a}
    % \end{subfigure}
    % \begin{subfigure}{.235\textwidth}
    %     \centering    
    %     \includegraphics[width=.98\linewidth, height=3.2cm]{figure/vis_dota/face4.jpg}
    %     \caption{Athlete's faces}
    %     \label{fig:fig9_b}
    % \end{subfigure} \\
%     \begin{subfigure}{.235\textwidth}
%         \centering    \includegraphics[width=.98\linewidth,height=3.5cm]{figure/vis_dota/text1.png}
%         \caption{Words on bulletin board}
%         \label{fig:fig10_a}
%     \end{subfigure}
%     \begin{subfigure}{.235\textwidth}
%         \centering    \includegraphics[width=.98\linewidth,height=3.5cm]{figure/vis_dota/text3.png}
%         \caption{Text seen on the elevator}
%         \label{fig:fig10_b}
%     \end{subfigure}
%     \vspace{-10pt}
%     \caption{Detection results on ICDAR2015}
%     \label{fig:fddb_icdar_vis}
% \end{figure}

\textbf{Performances on Other Datasets.}
We further do experiments on ICDAR2015, and HRSC2016 as shown in Tab. \ref{tab:hrsc2016_icdar2015}. For ICDAR2015, there are rich existing methods such as R$^2$CNN, Deep direct regression ~\cite{he2017deep} and FOTS ~\cite{liu2018fots}, and the current state-of-art has reached 91.67\%. They all have a lot of text-based tricks, but we find that they are also not aware of the rotation sensitivity error. Therefore, we conduct some verification experiments based on $\ell_{mr}$ and eight-parameter regression method. Positive results are obtained for all validation experiments on both datasets. Our detector performs competitively which shows the generalization of our method on scene text data. Besides, our method has also been verified on HRSC2016, and the experimental results are also comparable to state-of-art.

\begin{table}[!tp]
        \centering
        \resizebox{0.35\textwidth}{!}{
            \begin{tabular}{l|ccl|c}
                
                Method &   Plane && Car\quad\quad& mAP \\
                \shline
                YOLOv2 \cite{redmon2016you}  & 96.60 && 79.20 \quad\quad& 87.90 \\
                R-DFPN \cite{R21_yang2018automatic}  & 95.90 && 82.50 \quad\quad& 89.20 \\
                DRBox \cite{liu2017learning}  & 94.90& & 85.00 \quad\quad& 89.95 \\
                S$^2$ARN \cite{bao2019single}  & 97.60 && 92.20 \quad\quad& 94.90 \\
                RetinaNet-H \cite{R20_Yang2019R3Det} & 97.34 && 93.60  \quad\quad& 95.47 \\
                ICN \cite{R27_azimi2018towards} & - && -  \quad\quad & 95.67\\
                FADet \cite{li2019feature}  & 98.69 && 92.72 \quad\quad& 95.71 \\
                R$^3$Det \cite{R20_Yang2019R3Det}  & 98.20 && 94.14 \quad\quad& 96.17 \\
                \hline
                Ours (RSDet)   & \textbf{98.04} && \textbf{94.97} \quad\quad& \textbf{96.50}\\
        \end{tabular}}
        \vspace{-5pt}
        \caption{Performance evaluation on UCAS-AOD dataset.}
        \label{tab:ucas-aod}
        \vspace{-15pt}
\end{table}

\subsection{Overall Evaluation}
The results on DOTA are shown in Table \ref{tab:dota_sota}. The compared methods include i) traditional deep learning methods, such as Faster RCNN \cite{R16_Ren2015Faster} and RetinaNet \cite{R25_Lin2017Focal}; ii) scene text detection methods, like R$^2$CNN \cite{R22_Jiang2017R2CNN} and RRPN \cite{R30_ma2018arbitrary}; iii) recently published methods for multi-category rotation detectors, includes ICN \cite{R27_azimi2018towards}, RoI Transformer \cite{R29_ding2018learning}, SCRDet \cite{R28_Yang2018SCRDet} and R$^3$Det \cite{R20_Yang2019R3Det}. The results of DOTA reported here are all obtained by submitting predictions to official DOTA evaluation server. None of the compared methods pays attention to the problem of rotation sensitivity error. To make the comparison fair, clean and direct, we do not use multi-scale training and testing, oversized backbones, and model integration, which are often used on DOTA's leaderboard methods. For the overall mAP, our method's performance is 1.3\% higher than the existing best method (R$^3$Det+ResNet152 \cite{R20_Yang2019R3Det}). Tab. \ref{tab:ucas-aod} gives the comparison on UCAS-AOD dataset, where our method achieves 96.50\% for OBB task which outperforms all the published methods. Moreover, the amount of parameters and calculations added by our techniques are almost negligible, and they can be applied to all region based rotation detection algorithms. Visualization results on aerial images and natural images are shown in Fig. \ref{fig:dota_vis}.

\section{Conclusion}
In this paper, the issue of rotation sensitivity error (RSE) is formally identified and formulated for region-based rotated object detectors. RSE mainly refers to the loss discontinuity and the five-parameter regression inconsistency. We propose a novel modulated rotation loss $\ell_{mr}$ to address the loss discontinuity and optimize the regression inconsistency with the eight-parameter regression. As a result, the new detector termed as RSDet can be trained end-to-end. Extensive experiments demonstrate that RSDet achieves the state-of-art performance on the DOTA benchmark and is also proven good generalization and robustness on different datasets and multiple detectors.

{\small
\bibliographystyle{ieeetr}
\bibliography{egbib}
}

\end{document}